\documentclass[conference]{IEEEtran}

\usepackage{cite}

\ifCLASSINFOpdf
  
   \usepackage[pdftex]{graphicx}
  
\else
\fi
\usepackage{url}

\begin{document}
%
\title{Data Innovation for International Development: An overview of natural language processing for qualitative data analysis}

\author{\IEEEauthorblockN{Philipp Broniecki}
\IEEEauthorblockA{School of Public Policy\\
University College London\\
Email: philipp.broniecki.14@ucl.ac.uk}
\and
\IEEEauthorblockN{Anna Hanchar}
\IEEEauthorblockA{The Data Atelier\\
London, UK\\
Email: anna.h@thedataatelier.com}
\and
\IEEEauthorblockN{Slava J. Mikhaylov}
\IEEEauthorblockA{Institute for Analytics and Data Science\\
Department of Government\\
University of Essex\\
Email: s.mikhaylov@essex.ac.uk}}


%


\maketitle

\begin{abstract}
Availability, collection and access to quantitative data, as well as its limitations, often make qualitative data the resource upon which development programs heavily rely. Both traditional interview data and social media analysis can provide rich contextual information and are essential for research, appraisal, monitoring and evaluation. These data may be difficult to process and analyze both systematically and at scale. This, in turn, limits the ability of timely data driven decision-making which is essential in fast evolving complex social systems. In this paper, we discuss the potential of using natural language processing to systematize analysis of qualitative data, and to inform quick decision-making in the development context. We illustrate this with interview data generated in a format of micro-narratives for the UNDP Fragments of Impact project.

\end{abstract}


%
\IEEEpeerreviewmaketitle

\section{Introduction}

Practitioners in the development sector have long recognized the potential of qualitative data to inform programming and gain a better understanding of values, behaviors and attitudes of people and communities affected by their efforts. Some organizations mainly rely on interview or focus group data, some also consider policy documents and reports, and others have started tapping into social media data. Regardless of where the data comes from, analyzing it in a systematic way to inform quick decision-making poses challenges, in terms of high costs, time, or expertise required. 

The application of natural language processing (NLP) and machine learning (ML) can make feasible the speedy analysis of qualitative data on a large scale. 

We start with a brief description of the main approaches to NLP and how they can be augmented by human coding. We then move on to the issue of working with multiple languages and different document formats. We then provide an overview of the recent application of these techniques in a United Nations Development Program (UNDP) study.

\section{Supervised and Unsupervised Learning}

There are two broad approaches to NLP - supervised learning and unsupervised learning \cite{james2013introduction}.
Supervised learning assumes that an outcome variable is known and an algorithm is used to predict the correct variable. Classifying email as spam based on how the user has classified previous mail is a classic example. In social science, we may want to predict voting behavior of a legislator with the goal of inferring ideological positions from such behavior. In development, interest may center around characteristics that predict successful completion of a training program based on a beneficiary's previous experience or demographic characteristics.

Supervised learning requires human coding - data must be read and labelled correctly. This can require substantial resources. At the same time, the advantage is that validation of a supervised learning result is relatively straightforward as it requires comparing prediction results with actual outcomes. Furthermore, there is no need to label all text documents (or interview data from each respondent) prior to analyzing them. Rather, a sufficiently large set of documents can be labelled to train an algorithm and then used to classify the remaining documents. 

In unsupervised learning, the outcome variable is unknown. The exercise is, therefore, of a more exploratory nature. Here the purpose is to reveal patterns in the data that allow us to distinguish distinct groups whose differences are small, while variations across groups are large. In a set of text documents, we may be interested in the main topics about which respondents are talking. In social sciences, we may look for groups of nations within the international system that use a similar language or that describe similar issues, like small-island states prioritizing climate change. Identifying such groups is often referred to as `dimension reduction' of data. 

Validation of unsupervised learning results is less straight-forward than with supervised learning. We use data external to our analysis to validate the findings \cite{Grimmer2013}. 

A complementary approach to unsupervised and supervised learning is the use of crowdsourced human coders. \cite{Benoit2016} show that crowdsourcing text analysis is a way to achieve reliable and replicable results quickly and inexpensively through the CrowdFlower platform. This approach can work well in supporting a supervised approach where outcomes need to be labelled. For example, \cite{Benoit2016} use this technique to produce party positions on the economic left-right and liberal-conservative dimensions from party manifestos. Online coders receive small specific tasks to reduce individual biases. Their individual responses are then aggregated to create an overall measure of party positions.

\section{Working with Multiple Languages}

A common obstacle to analyzing textual data in many fields, including the international development sector, is the plethora of languages that practitioners and researchers need to consider -- each with subtle but meaningful differences. Fortunately, significant commercial interest in being able to translate large quantities of text inexpensively has led to major advances in recent years driven by Microsoft, Google, and Yandex with the introduction of neural machine translation \cite{wu2016google}. They provide services that are free of charge that can be easily integrated into standard programming languages like Python and R.\footnote{Bing, Google, and Yandex place character limits on single document translations, with the latter allowing for slightly longer text chunks. For longer documents, commercial alternatives would have to be used.} Open source neural machine translation systems are also being made available \cite{klein2017opennmt}. 

In a recent application, \cite{benoit2012sincerity} estimate the policy preferences of Swiss legislators using debates in the federal parliament. With speeches delivered in multiple languages, the authors first translate from German, French, and Italian into English using Google Translate API. They then estimate the positions of legislators using common supervised learning methods from text and compare to estimates of positions from roll-call votes. 

\cite{de2017lost} evaluate the quality of automatic translation for social science research. The authors utilize the \emph{europarl} dataset \cite{koehn2005europarl} of debate transcripts in the European Parliament and compare English, Danish, German, Spanish, French, and Polish official versions of the debates with their translations performed using Google Translate. \cite{de2017lost} find that features identified from texts are very similar between automatically translated documents and official manual translations. Furthermore, topic model estimates are also similar across languages when comparing machine and human translations of EU Parliament debates.  

\section{Working with Documents}

In recent years, great strides have been made into leveraging information from text documents. For example, researchers have analyzed speeches, legislative bills, religious texts, press communications, newspaper articles, stakeholder consultations, policy documents, and regulations. Such documents often contain many different dimensions or aspects of information and it is usually impossible to manually process them for systematic analysis. The analytical methods used to research the content of such documents are similar. We introduce prominent applications from the social sciences to provide an intuition about what can be done with such data.

\subsection{Open-ended survey questions}

Open-ended questions are a rich source of information that should be leveraged to inform decision-making. We could be interested in several aspects of such a text document. One useful approach would be to find common, recurring topics across multiple respondents. This is an unsupervised learning task because we do not know what the topics are. Such models are known as topic models. They summarize multiple text documents into a number of common, semantic topics. \cite{Roberts2014} use a structural topic model (STM) that allows for the evaluation of the effects of structural covariates on topical structure, with the aim of analyzing several survey experiments and open-ended questions in the American National Election Study.

\subsection{Religious statements}

\cite{Lucas2015a} analyze Islamic fatwas to determine whether Jihadist clerics write about different topics to those by non-Jihadists. Using an STM model, they uncover fifteen topics within the collection of religious writings. The model successfully identifies characteristic words in each topic that are common within the topic but occur infrequently in the fourteen other topics. The fifteen topics are labelled manually where the labels are human interpretations of the common underlying meaning of the characteristic words within each topic. Some of the topics deal with fighting, excommunication, prayer, or Ramadan. While the topics are relatively distinct, there is some overlap, i.e. the distance between them varies. This information can be leveraged to map out rhetorical network.

\subsection{Public debates}

\cite{Blumenau2017} uses topic modeling to link the content of parliamentary speeches in the UK's House of Commons with the number of signatures of constituency-level petitions. He then investigates whether the signatures have any bearing on the responsiveness of representatives, i.e. whether Members of Parliament take up an issue if more people sign a petition on that issue. Also using speeches from the UK House of Commons, \cite{Blumenau2017a} produces evidence for the female role-model hypothesis. He shows that the appointment of female ministers leads to more speaking time and speech centrality of female backbenchers.

\subsection{News reports}

\cite{smidt2017} uses supervised machine learning to generate data on UN peacekeeping activities in Cote d'Ivoire. The text data inputs are news articles from the website of the UN peacekeeping mission in Cote d'Ivoire. Based on a manually classified subset of articles, an algorithm is trained to classify terms into activity categories. Based on this algorithm, the remaining articles are then categorized. Analyzing these data yields new micro-level insights into the activities of peacekeepers on the ground and their effects.

\cite{Grimmer2012} take a similar approach to researching the effectiveness of self-promotion strategies of politicians. They analyze 170,000 press releases from the U.S. House of Representatives. First, 500 documents are classified by hand into five categories of credit claiming, next the supervised learning algorithm, ReadMe \cite{Hopkins2010}, is used to code the remaining documents automatically. Using this data, they show that the number of times legislators claim credit generates more support than whether or not the subject they claim credit for amounts to much. 

\subsection{Sentiment analysis}

Instead of uncovering topics, we may want to know of a positive or negative tone of any given document. In an open-ended survey response, we could be interested in how the respondent rates the experience with the program. Sentiment analysis is a common tool for such a task. It is based on dictionaries of words that are associated with positive or negative emotions. The sentiment of a document such as an open-ended question would then be based on relative word counts and the sentiment scores with which these words are associated. \cite{johnston2007chinese} find positive and negative keywords and count their frequency in Chinese newspaper articles that mention the United States. With this data, they identify attitudes towards the United States in China. 

\subsection{Text reuse}

Another application is to study text reuse in order to trace the flow of ideas. \cite{Wilkerson2015} analyze bill sections to identify whether two sections of a bill propose similar ideas. The algorithm they use was devised to trance gene sequencing and takes the frequency and order of words into account. Using this technique, they can measure the influence of one bill on another. Similarly, \cite{Hinkle2015} studies policy diffusion by shedding light on how the American Supreme Court and Courts of Appeals influence diffusion of state policies. He uses plagiarism software to quantify the exact degree to which an existing law is reflected in a new proposal. Tracing ideas or influence over time and space can generate insights into the sources of information, the degree of spillover, and the influence of certain actors, ideas, or policies. It can shed light on network structures and long-term effects that would otherwise be hidden to us.

\subsection{Estimating preferences of actors}

In the social sciences, text is often used to infer preferences. Various scaling techniques have been developed and refined over recent years. Wordfish is a scaling algorithm that enables us to estimate policy positions based on word frequencies \cite{Slapin2008}. Researchers have used this approach to measure policy positions on European integration in the European Parliament \cite{Slapin2010}, on austerity preferences in Ireland \cite{Herzog2015}, and intra-party preferences in the energy debate in Switzerland \cite{Schwarz2015}. Recent developments in the field allow researchers to estimate attitudes in multiple issue dimensions and, therefore, allow for more fine-grained preference estimates \cite{Lauderdale2016}. 

\subsection{Taking context into account}

More recent developments in NLP depart from frequencies of single words or groupings of multiple words. Instead, each word is an observation and its variables are other words or characters. Thus, each word is represented by a vector that describes words and their frequencies in the neighborhood. This approach allows for the capture of text semantics \cite{Mikolov}. 

\cite{Shahbazi2016} apply this to evaluating real estate in the U.S. by comparing property descriptions with words that are associated with high quality. \cite{Baturo2017} collected all country statements made during the United Nations General Debate where heads of state and government make statements that they consider to be important for their countries. Using this data, \cite{Gurciullo2016} run a neural network. They construct an index of similarity between nations and policy themes that allows us to identify preference alliances. This enables them to identify major players using network centrality and show that speeches contain information that goes beyond mere voting records.

In a large exploratory effort, \cite{Gurciullo} use dynamic topic modeling which captures the evolution of topics over time, along with topological analytical techniques that allow for the analysis of multidimensional data in a scale invariant way. This enables them to understand political institutions, in this case through the use of speeches from the UK House of Commons. They classify representatives into groups according to speech content and verbosity, and identify a general pattern of political cohesion. They further show that it is feasible to track the performance of politicians with regard to specific issues using text. Topological techniques are especially useful to discover networks of relations using text. \cite{Gurciullo2016} apply this to uncover ideological communities in the network of states in the international system using UN General Assembly speeches.

\section{Working with Short Text, Micro-Blogs, Social Media}

Social media networks such as Twitter, the microblogging service, or the social network, Facebook, connect a vast amount of people in most societies. They generally contain shorter text excerpts compared to the sources of text previously discussed. However, their size and dynamic nature make them a compelling source of information. Furthermore, social networks online reflect social networks offline \cite{Bisbee2017}. They provide a rare and cheap source of information on dynamic micro-level processes.

\subsection{Twitter}

Similar to our discussion above, topic models can be used to analyze social media data. \cite{Lucas2015a} use such a model to analyze how the United States is viewed in China and in Arabic-speaking countries in response to the disclosure of classified information by Edward Snowden. They collect tweets containing the word ``Snowden'' in both languages. The tweets are then translated to English using machine translation. \cite{Lucas2015a} show that Chinese posts are concerned more about attacks in terms of spying, while Arabic posts discuss human rights violations.

We can use social media to analyze networks and sentiments. Similar to word counts, volume of posts can carry information. \cite{Nulty2016} collect tweets originating from and referring to political actors around the 2014 elections to the European Parliament. They consider the language and national distribution as well as the dynamics of social media usage. Using network graphs depicting the conversations within and between countries, they identify topics debated nationally, and also find evidence for a Europe-wide debate around the EP elections and the European Union generally. Using sentiment analysis, they further show that positive statements were correlated with pro-integration attitudes whereas negative debates were more national and anti-integration. 

This EU example translates well to national conversations involving multiple ethnic or linguistic groups elsewhere. Moreover, we can learn how information spreads from social networks. Consequently, within ethical boundaries, we may also be able to target information more efficiently. An analysis of Twitter data from the Arab Spring suggests that coordination that originated from the periphery of a network rather than the center sparked more protest \cite{Steinert-Threlkeld2017}. Coordination was measured as a Gini index of Hashtags while centrality was measured by a count of followers of an account. 

\subsection{Facebook}

Social media has been used to estimate preferences as well. The advantage of social media compared to speeches or any other preference indicator is coverage. \cite{Bond2015} use endorsement of official pages on Facebook to scale ideological positions of politicians from different levels of government and the public into a common space. Their method extends to other social media such as Twitter where endorsements and likes could be leveraged.

\subsection{Weibo, RenRen, and Chinese microblogs}

The most prominent example of supervised classification with social media data involves the first large scale study of censorship in China. \cite{King2013} automatically downloaded Chinese blogposts as they appeared online. Later they returned to the same posts and checked whether or not they had been censored. Furthermore, they analyzed the content of the blog posts and showed that rather than banning critique directed at the government, censorship efforts concentrate on calls for collective expression, such as demonstrations.

Further investigations of Chinese censorship were made possible by leaked correspondence from the Chinese Zhanggong District. The leaks are emails in which individuals claim credit for propaganda posts in the name of the regime. The emails contain social media posts and account names. \cite{King2017} used the leaked posts as training data for a classification algorithm that subsequently helped them to identify more propaganda posts. In conjunction with a follow-up survey experiment they found that most content constitutes cheerleading for the regime rather than, for example, critique of foreign governments.

In the next section we discuss an application of natural language processing in international development research.

\section{UNDP Fragments of Impact Initiative}

In 2015, the United Nations Development Programme (UNDP) Regional Hub for Europe and CIS launched a Fragments of Impact Initiative (FoI) that helped to collect qualitative (micro-narratives) and quantitative data from multiple countries. 

Within a six-month period, around 10,000 interviews were conducted in multiple languages. These covered the perception of the local population in countries including Tajikistan, Yemen, Serbia, Kyrgyzstan and Moldova on peace and reconciliation, local and rural development, value chain, female entrepreneurship and empowerment, and youth unemployment issues. The micro-narratives were collected using SenseMaker(r), a commercial tool for collecting qualitative and quantitative data. The micro-narratives were individual responses to context-tailored questions. An example of such a question is: ``Share a recent example of an event that made it easier or harder to support how your family lives.''

While the analysis and visualization of quantitative data was not problematic, systematic analysis and visualization of qualitative data, collected in a format of micro-narratives, would have been impossible. 

To find a way to deal with the expensive body of micro-narrative data, UNDP engaged a group of students from the School of Public Policy, University College London, under the supervision of Prof Slava Mikhaylov (University of Essex) and research coordination of Dr Anna Hanchar (The Data Atelier). The objective of this work was to explore how to systematize the analysis of country-level qualitative data, visualize the data, and inform quick decision-making and timely experiment design. The results of the project were presented at the Data for Policy 2016 \cite{insights}.

The UCL team had access to micro-narratives, as well as context specific meta-data such as demographic information and project details. For a cross-national comparison for policy-makers, the team translated the responses in multiple languages into English using machine translation, in this case Translate API (Yandex Technologies). As a pre-processing step, words without functional meaning (e.g. `I'), rare words that occurred in only one narrative, numbers, and punctuation were all removed. The remaining words were stemmed to remove plural forms of nouns or conjugations of verbs. 

\begin{figure}
\centering

\includegraphics[width=.43\textwidth]{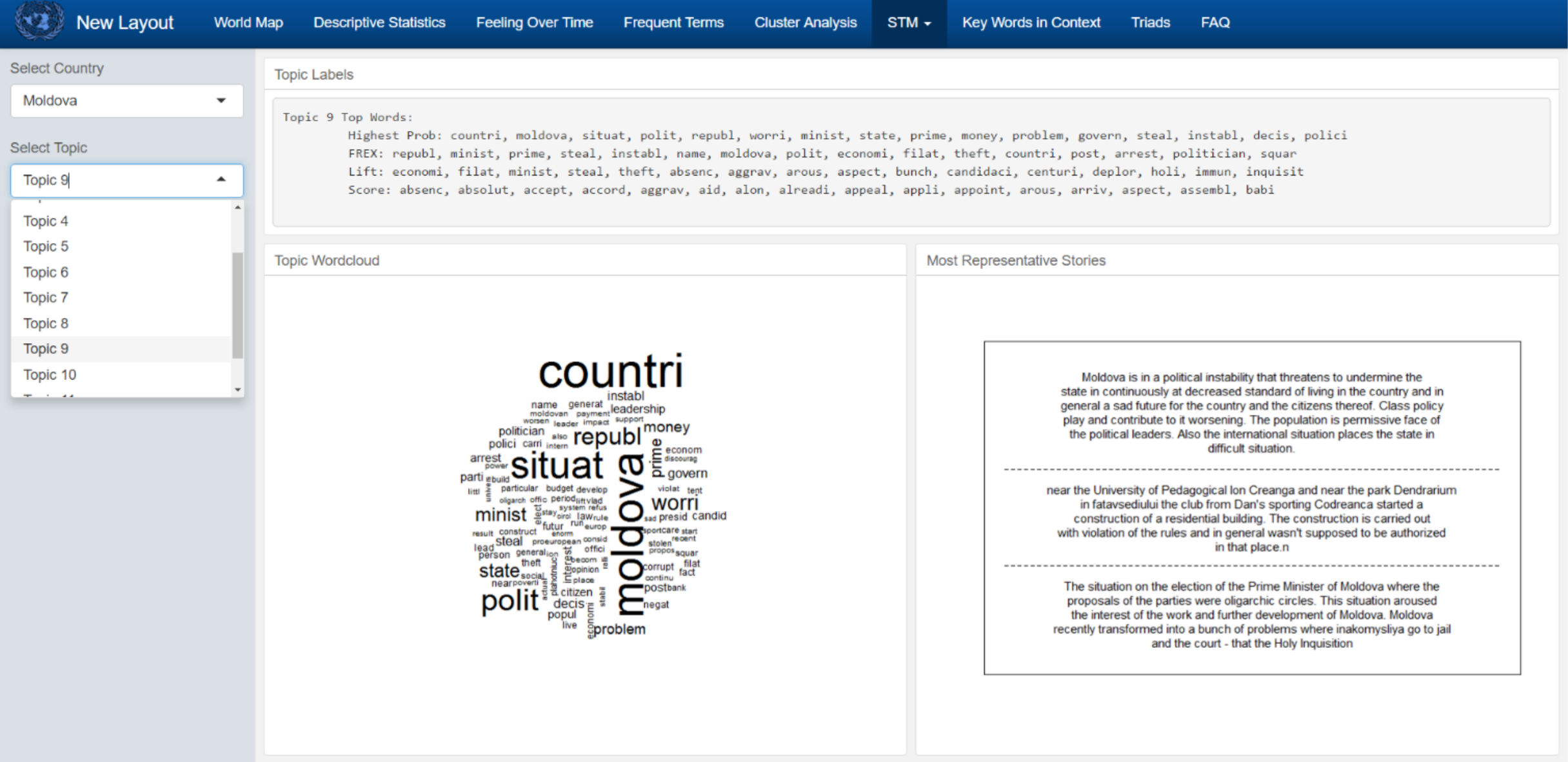} 
\caption{\emph{Dashboard Interface} Example of topic modeling results for Moldova showing the highest probability words for one of the topics and corresponding most representative micro-narratives.  
\label{fig:fig1}}
\end{figure}

As part of this exploration exercise, and guided by UNDP country project leads, the UCL team applied structural topic modeling \cite{Roberts2014} as an NLP approach and created an online dashboard containing data visualization per country. The dashboard included descriptive data, as well as results. Figure \ref{fig:fig1} illustrates an example of the dashboard. The analysis also allowed for the extraction of general themes described by respondents in the micro-narratives, and looked for predictors such as demographics that correlated with these themes. In Moldova, the major topic among men was rising energy prices. Among women the main topic was political participation and protest, which suggests that female empowerment programs could potentially be fruitful. 

\begin{figure}
\centering

\includegraphics[width=.45\textwidth]{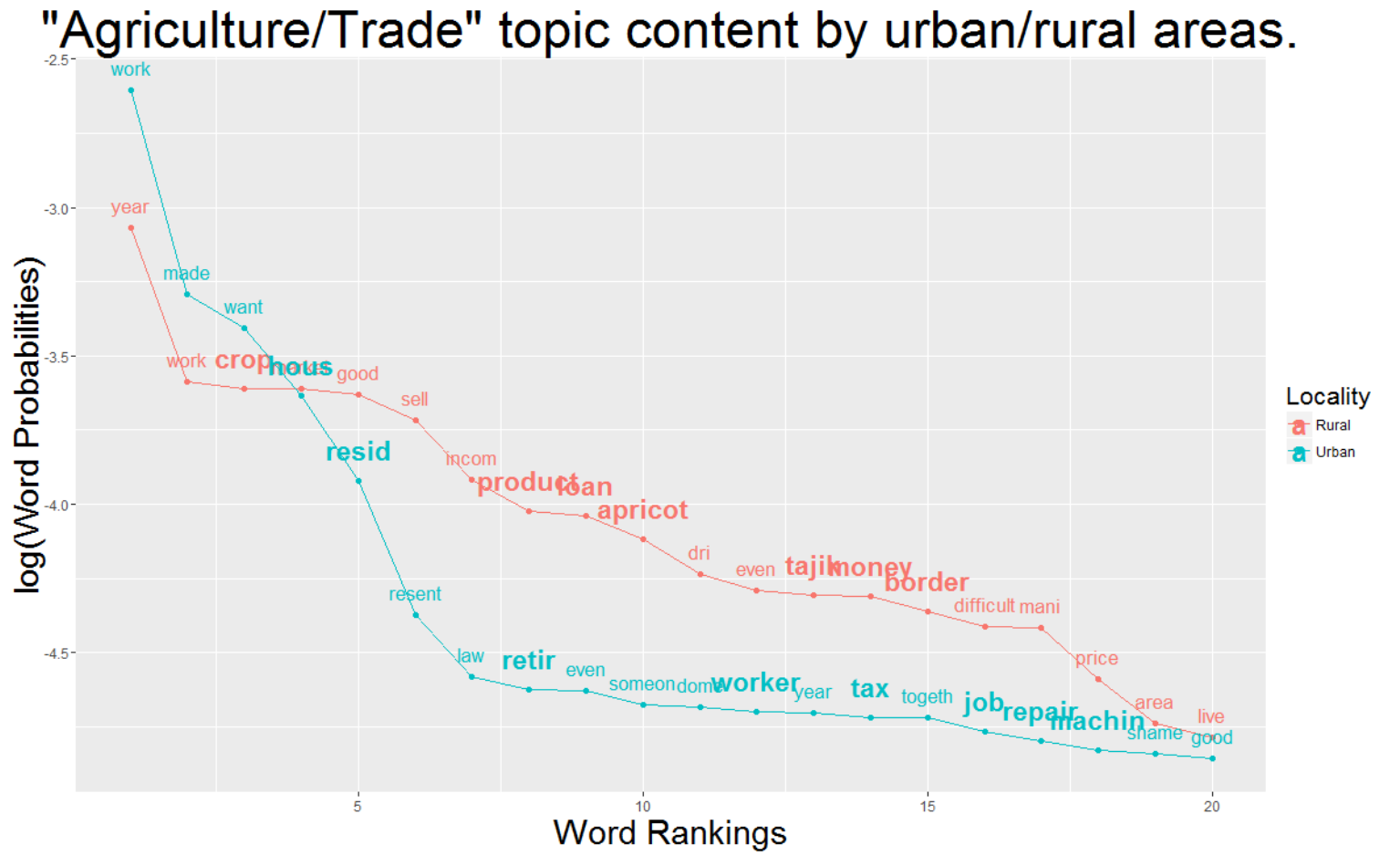} 
\caption{\emph{Comparing topic prevalence and topical content for urban and rural areas in Kyrgyzstan}.  
\label{fig:fig2}}
\end{figure}

In Kyrgyzstan, the team found that the main topics revolved around finding work, access to resources and national borders. Using the meta-data on urbanization, it became clear that rural respondents described losing livestock that had crossed the border to Tajikistan, or that water sources were located across the border. The urban population was concerned about being able to cross the border to Russia for work. Figure \ref{fig:fig2} shows word probabilities from the main ``agriculture/trade'' topic across respondents from urban and rural communities. 

\begin{figure}
\centering

\includegraphics[width=.45\textwidth]{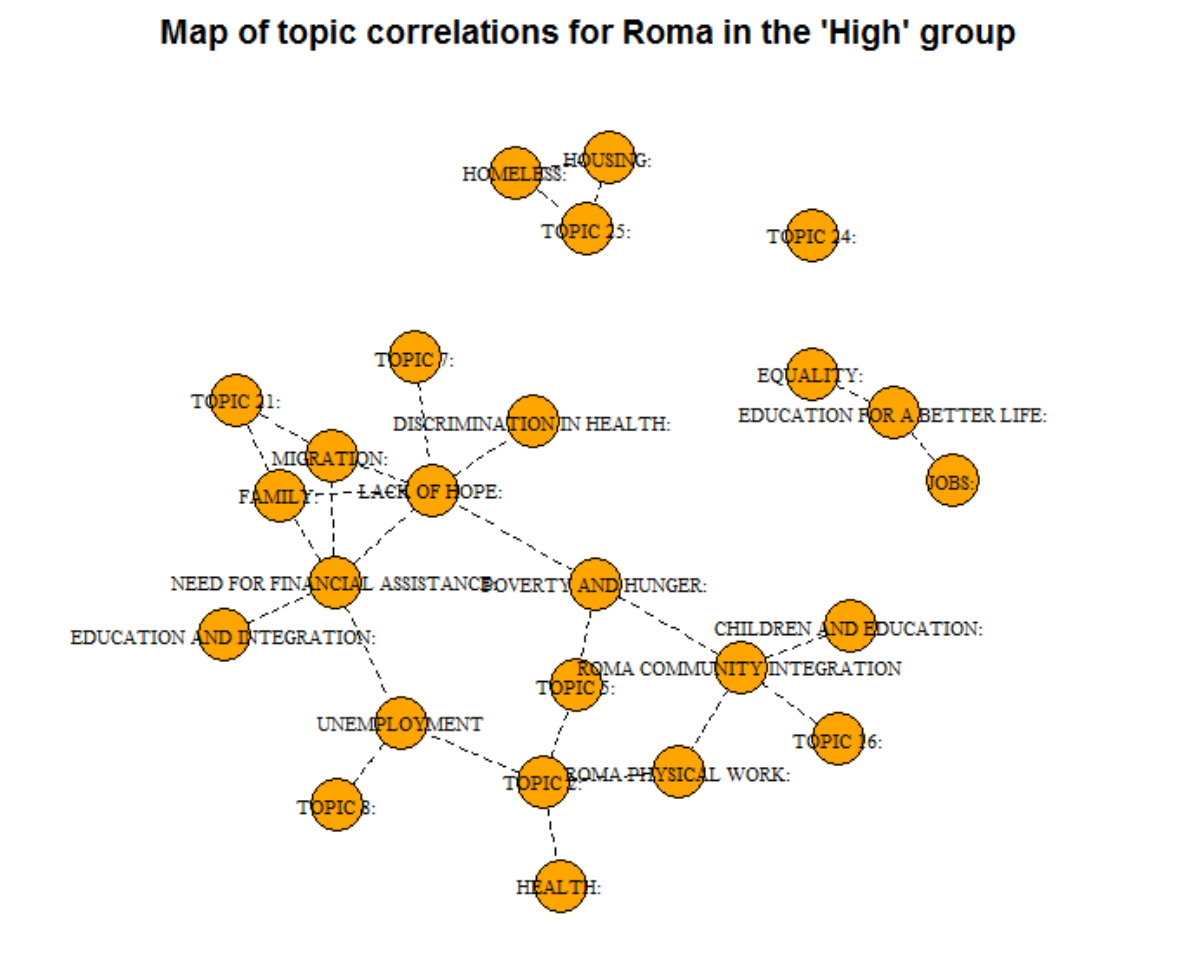} 
\caption{\emph{Topic correlations in areas of high Roma concentration in Serbia.}  
\label{fig:fig3}}
\end{figure}

For Serbia, the analysis compared issues faced by Roma populations in areas of high and low Roma concentration. Figure \ref{fig:fig3} shows the relationship between topics discussed by Roma respondents in areas of high concentration. 

 \cite{insights} found that Roma respondents identified education as the overarching main topic, independent of the density of the Roma population. Differences were found between respondents across the level of integration with society. In areas of high Roma concentration, respondents were aware of available channels for inclusion. In low Roma density areas, respondents were mainly concerned with severe poverty and discrimination preventing societal inclusion. 

\begin{figure}
\centering

\includegraphics[width=.4\textwidth]{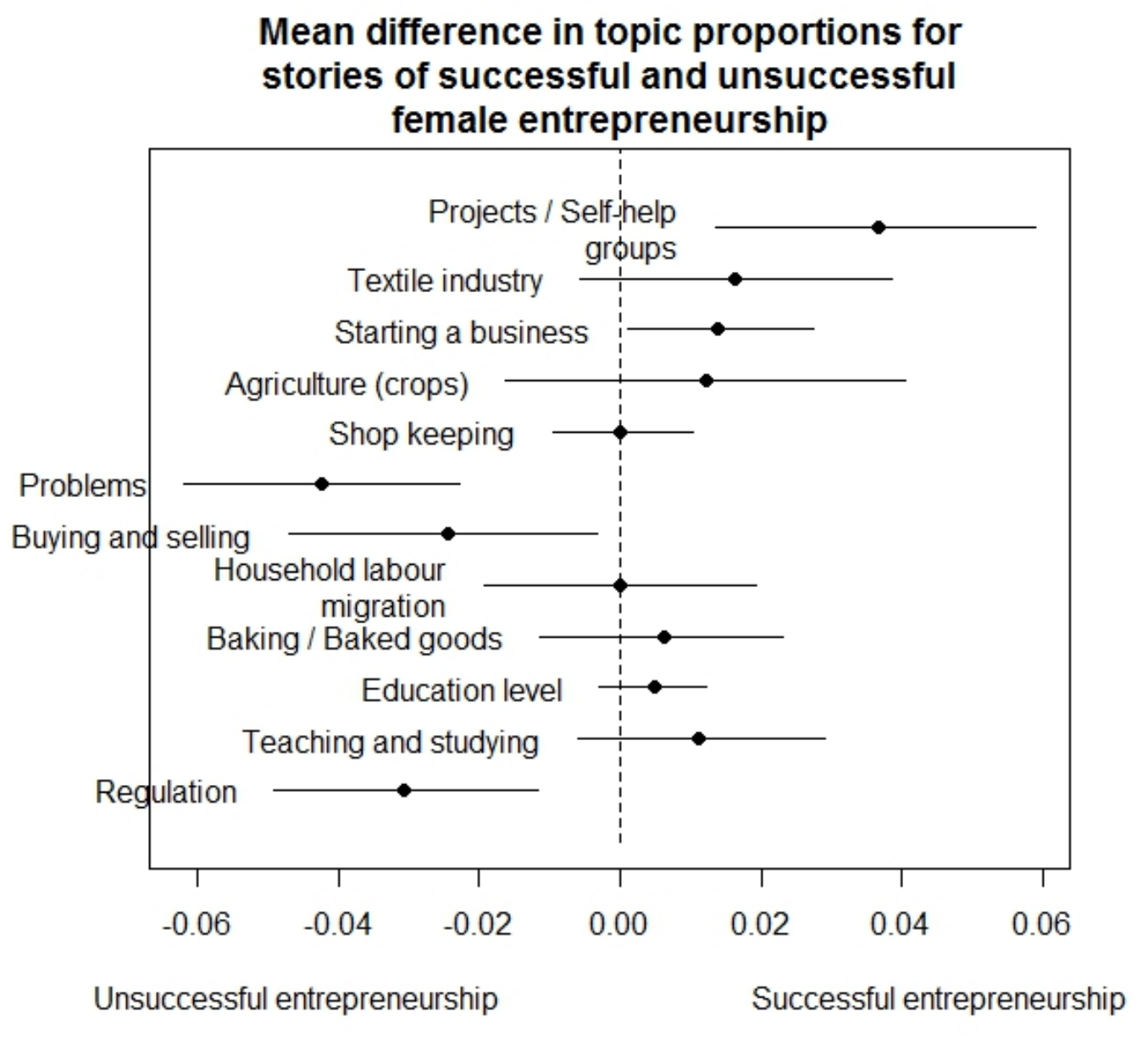} 
\caption{\emph{Identifying constraints on female entrepreneurship in Tajikistan}  
\label{fig:fig4}}
\end{figure}

In Tajikistan, \cite{insights} investigated the relationship between household labor migration and female entrepreneurship success. They found strong regional differences between the Sughd and Khatlon regions where topics in less successful Khatlon revolved around red tape. Moreover, successful entrepreneurship was very much related to receiving remittances. Figure \ref{fig:fig4} illustrates topics that correlate with success.

\begin{figure}
\centering

\includegraphics[width=.45\textwidth]{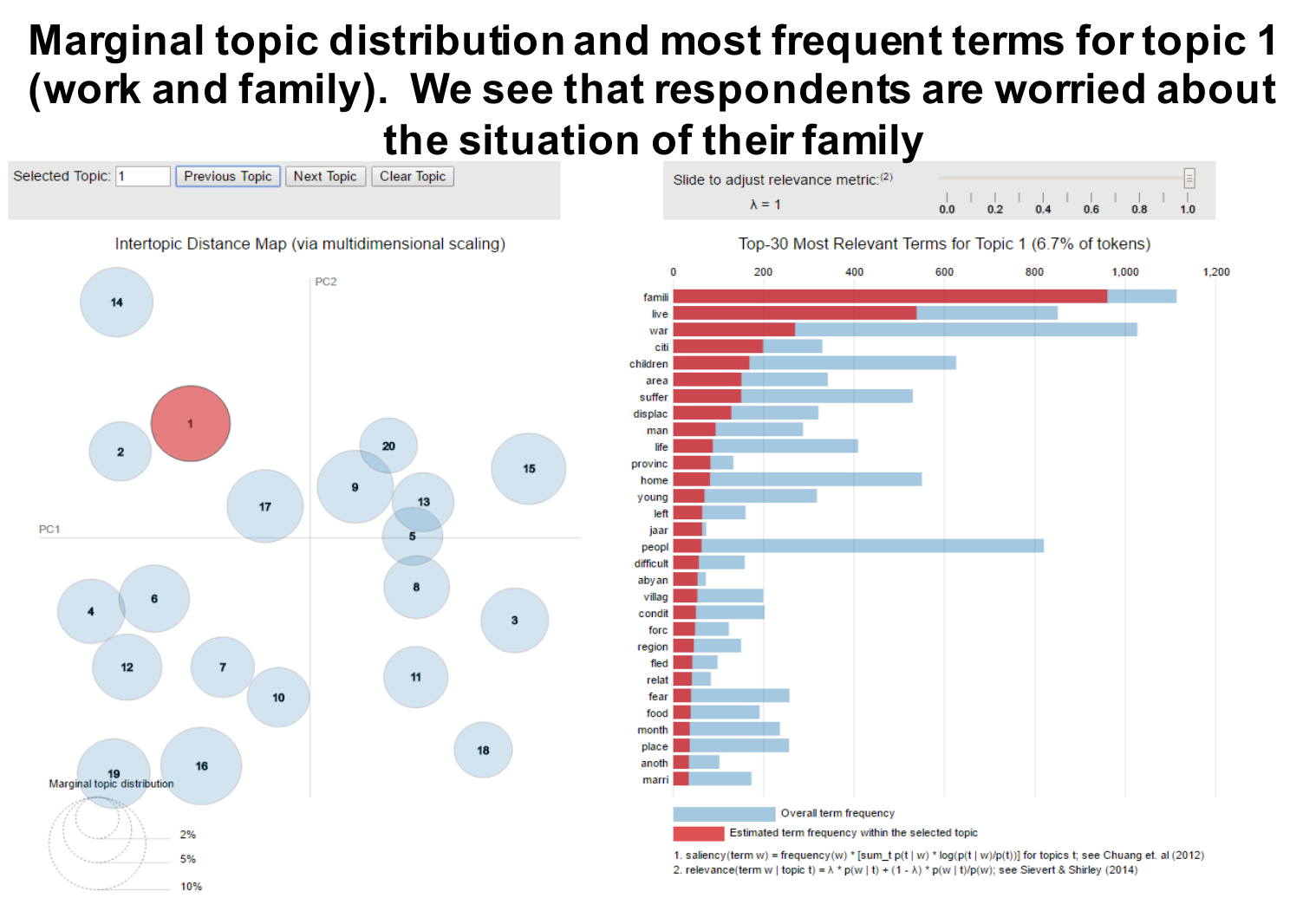} 
\caption{\emph{Exploring differences in perspectives by age in Yemen}  
\label{fig:fig5}}
\end{figure}

Analysis of micro-narratives from Yemen showed that the most recurrent themes focused on family issues (Figure \ref{fig:fig5}). There are significant differences in terms of engagement in civil society between young people and the older population. Young respondents emphasized pro-active behavior, political engagement, and interest in community-driven initiatives fostering political development. 

\section{Conclusion}

In this overview, our aim has been to demonstrate how new forms of data can be leveraged to complement the work of practitioners in international development. We have demonstrated that a wide variety of questions can be asked. 

Exploratory work can be performed to systematize large quantities of text. Additionally, we can learn about the sentiment that specific groups express towards specific topics. Networks can be uncovered, the spread of information or ideas can be traced, and influential actors identified. We can classify documents based on human coding of a subset of documents and establish which topics predict/correlate with predefined outcomes such as successful employment or completion of a program.

While the application used here to illustrate the discussion focuses on texts in the form of open-ended questions, social networks can be used and their coverage and topicality can be leveraged.

Natural language processing has the potential to unlock large quantities of untapped knowledge that could enhance our understanding of micro-level processes and enable us to make better context-tailored decisions.

\section*{Acknowledgment}

Authors' names are listed in alphabetical order. Authors have contributed equally to all work.




\bibliographystyle{IEEEtran}
\bibliography{IEEEabrv,library}

%
%
%

\end{document}